# Dermatologist vs Neural Network

Kaushil Mangaroliya[1*]                                Mitt Shah[1*]

18bce091@nirmauni.ac.in                          18bce121@nirmauni.ac.in

[1.] Institute of Technology, Nirma University, India

## Abstract

*Cancer, in general, is very deadly. Timely treatment of any cancer is the key to saving a life. Skin cancer is no exception. There have been thousands of Skin Cancer cases registered per year all over the world. There have been 123,000 deadly melanoma cases detected in a single year. This huge number is proven to be a cause of a high amount of UV rays present in the sunlight due to the degradation of the Ozone layer. If not detected at an early stage, skin cancer can lead to the death of the patient. Unavailability of proper resources such as expert dermatologists, state of the art testing facilities, and quick biopsy results have led researchers to develop a technology that can solve the above problem. Deep Learning is one such method that has offered extraordinary results. The Convolutional Neural Network proposed in this study out performs every pretrained models. We trained our model on the HAM10000 dataset which offers 10015 images belonging to 7 classes of skin disease. The model we proposed gave an accuracy of 89%. This model can predict deadly melanoma skin cancer with a great accuracy. Hopefully, this study can help save people's life where there is the unavailability of proper dermatological resources by bridging the gap using our proposed study.*

**Key Words** – Skin Cancer, HAM10000, Convolutional Neural Network, Melanoma, Dermatologist,

## 1. Introduction

Over the time number of people suffering from skin problems have been increased, especially the number of skin cancer cases. Skin cancer is becoming a global health issue mainly due to the sunlight carrying harmful UV- rays [12]. As of now, in 1 year, we are treating over 123,000 melanoma and 30,00,000 non-melanoma cases [1]-[5]. Each year addition of 4,500 melanoma cases and the addition of 300,000 non-melanoma cases have been observed to the existing number due to around 10% depletion of the ozone layer around the earth [1].

The only way a patient can be saved from death is the early detection of skin cancer so doctors can act quickly. Seeing the severity of an increasing number of cases and mortality rates due to this deadly cancer, there have been ideas like using machine learning and deep learning to detect skin cancers at an earlier stage. Esteva et. al, [6], got great breakthrough results and inspired many research papers since then. Many researchers have been giving their models on a variety of skin cancer classification tasks [6-11]. The current approach to handle this problem revolves around transfer learning. Now that large models such as ResNet [13], AlexNet [14], VGG [15] have been developed, researchers tend to use these models to approach this problem. Many researchers have used models like ResNet, MobileNet, Inception, etc. directly to solve a serious problem like this.

There is much decent work but these classification lack general capability [16][17][18] because of their not so satisfying accuracy and results when it comes to the multi-class classification of skin cancer [7][19][20]. This paper tries to overcome that shortcoming by using CNN and 10015 dermoscopic images to train the CNN we made. These images come from a dataset called HAM10000 Dataset[21][22]. The dataset used has 7 different classes of skin cancer images in it. The model we created is capable of classifying the images into all 7 skin cancer classes. To understand skin cancer and its behavior we have also conducted some data analysis to give more insight.

## 2. Dataset

The dataset used here is HAM10000 Dataset [38]. A part of it is used for training and a small portion for validation and testing the model created. The dataset comprises of total 10015 images of 600 X 450 resolution. It comprises 7 classes of skin cancer namely Melanocytic nevi, Melanoma, Benign keratosis, Basal cell carcinoma, Actinic keratosis, Vascular, and Dermatofibroma skin cancer types. The images provided in the dataset as it is cannot be used to train the model directly so we need to do some pre-processing. Figure 1 shows some images of skin cancer.

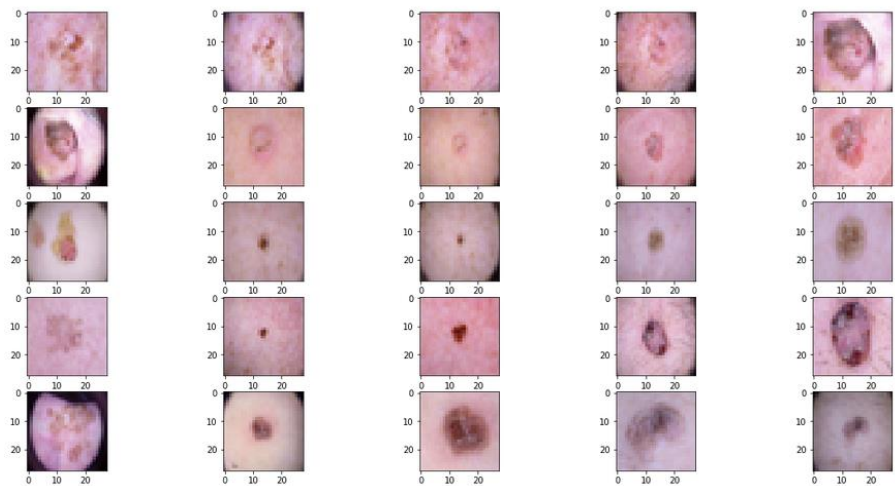

*Figure 1. Sample images from HAM10000 dataset representing dermoscopic images of cancerous skin.*

## 2.1 Data Analysis

To understand the behavior of skin cancer properly and gain insights we need to study the dataset completely. First, we need to check for class-imbalance because in any dataset number of instances in any class is never going to be the same. Figure 2 shows a number of data points of each skin cancer class and it is observed that Melanocytic Nevi comprises more than 65% of the overall dataset, this will have an impact later during the classification so we need to deal with it.

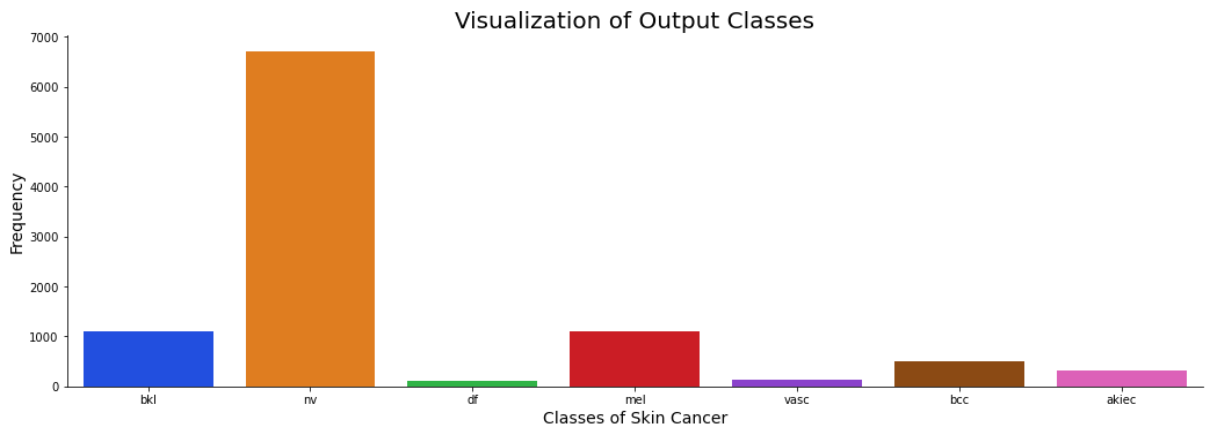

*Figure 2. Shows how many patients of each skin cancer type are there in dataset.*

Skin cancer is conformed via Histopathology, to see if this is really the case for every patient. Figure 3 shows that all of the Melanocytic nevi are conformed through follow-up, this is a strange behavior and it is not seen for other skin cancer classes.

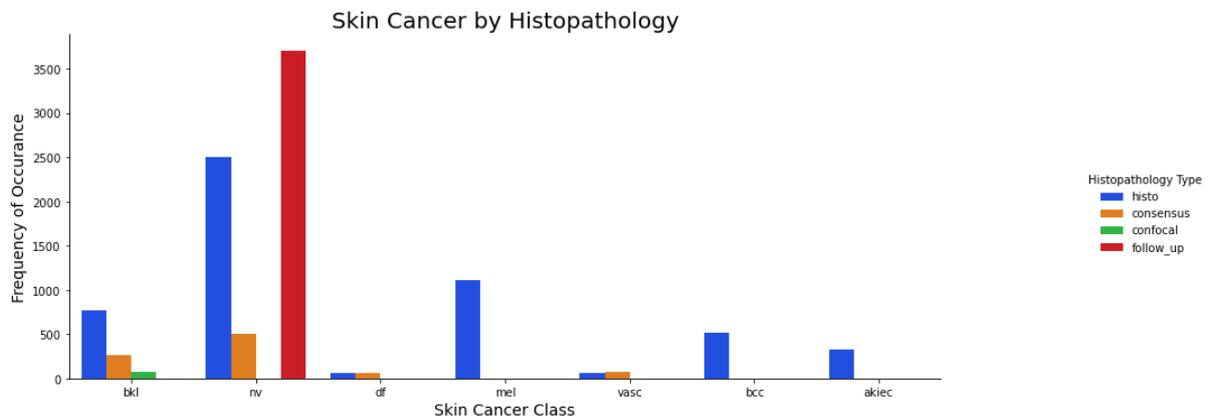

*Figure 3. Shows what type of procedure was followed to conform the skin cancer.*

While studying the location of skin cancer on the body, the cause of skin cancer can be understood. Figure 4 shows skin cancer by localization. Skin cancer seems to have more occurrences in the back and the lower extremity of people. This may be due to due to over-exposure to harmful sunlight carrying ultra-violet rays.

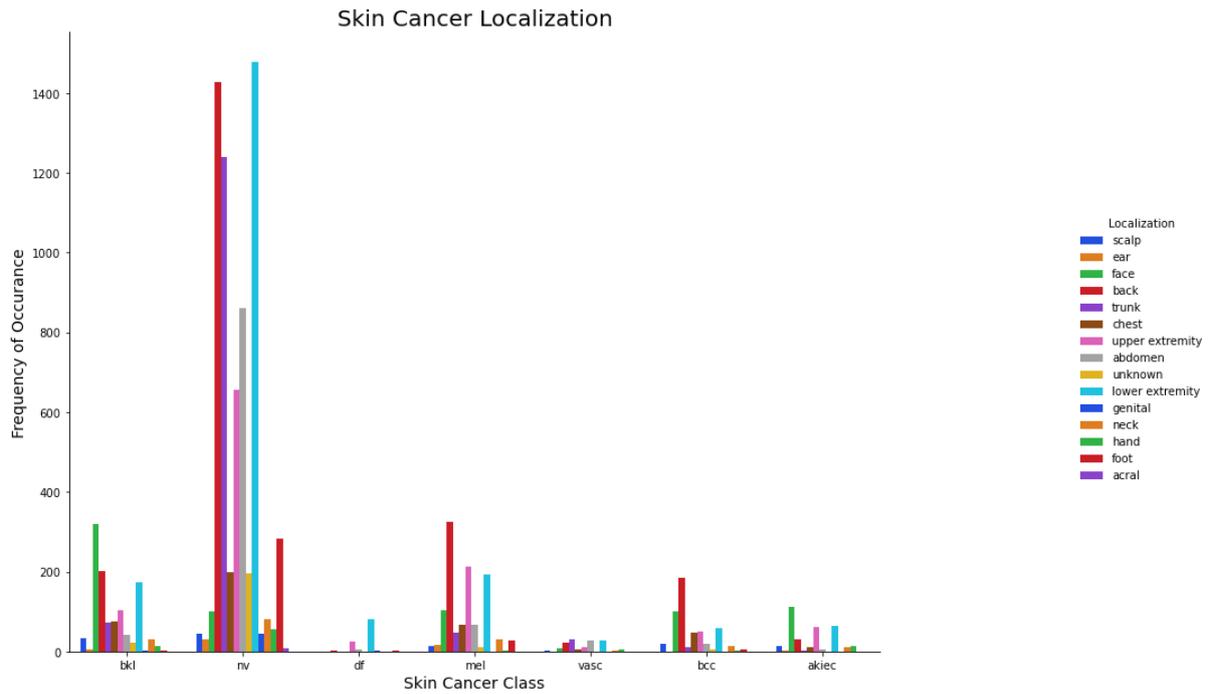

*Figure 4. Shows where the infected skin is present on body of patient's body.*

To understand who are these people getting affected by skin cancer, children, middle-aged people or elderly people is very important. Figure 5 shows that skin cancer is dominant among middle-aged people and not observed so frequently in children or elderly people.

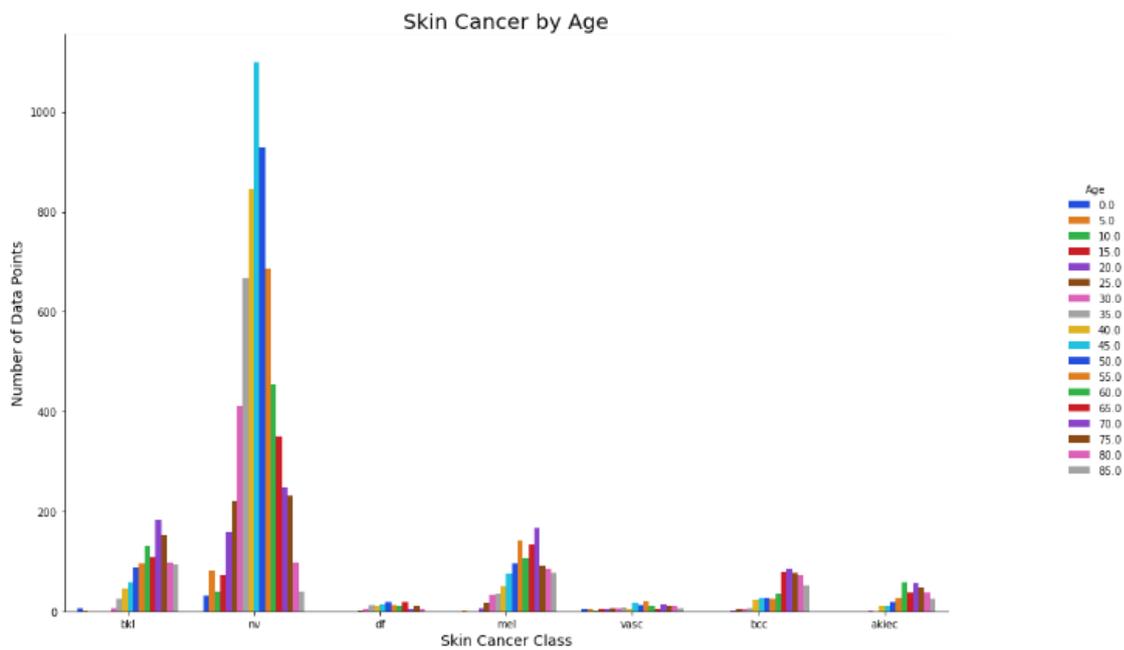

*Figure 5. Shows what is the age group of particular patients with particular skin cancer type*

## 2.2 Data pre-processing

The first thing to check for is all the labels and null values if any or error in labeling. Seeing the dataset there is no image that stands unlabeled. We resized the images to dimensions 28x28. The dataset of 10015 images is to be divided into 2 partitions, training set, and validation set. To maintain the authenticity of validation, we need to make sure that there are no images from the training set present in the validation set. For this particular model, we take 9013 images as the training set and 1002 images as the validation set. This is a typical 90% to 10% split ratio for training a model.

## 2.3 Data Augmentation

Data Augmentation is an effective technique for overcoming class imbalance. Data Augmentation is a strategy by which we can increase the variance in the dataset without actually collecting new images. This dataset suffers from a class imbalance problem which might affect the F1 score at a later stage. We made multiple changes to the images in many ways possible. Images were normalized, were flipped horizontally and vertically, rotated by a maximum of 15 degrees, changed the brightness, zoomed to the central portion. This helped in improving the accuracy by a significant margin. As Data Augmentation increases the size of the dataset hence it also helps in reducing overfitting.

# 3. Model Architecture

Since the past decade, there has been a significant advancement in knowledge related to Convolutional Neural Networks ( CNN ). CNN has been widely used to achieve different goals such as image classification, object detection, instance segmentation, and the list goes on. While dealing with images, CNN is far more efficient than a traditional Dense Neural Network. When traditional Dense Neural Network is used with image pixels, the number of parameters increases drastically which leads to the necessity of high computational power which on the other hand is not the case with CNN. CNN is a type of Neural Network that takes an image as its input. As an output, we get a class to which the image belongs to or what all objects are present in that particular image.

In our case, we will be using a CNN which will be trained on thousands of images of skin cancer and as a result, the model will be able to classify an unknown image into one of the 7 classes of skin cancer.

There has been extensive use of Transfer Learning by researchers to classify images without a lot of hard work. To achieve the desired accuracy for skin cancer classification, instead of using Transfer Learning we have built our own CNN which is better or as good as the previous Transfer Learning models proposed by other researchers. In this model, we have used 20 layers to build our CNN. These 20 layers include a bunch of Convolutional layers, Max Pooling layers, Dropout layers, and Dense layers. Figure 6 shows all details like layer with its shape and number of parameters, layer-wise, and total.

Transfer learning is the process in which a pre-trained Neural Network is used. This pre-trained network is then changed slightly according to our needs for a particular problem. Some examples of pre-trained Neural Networks are ResNet50, InceptionV3, Vgg19, MobileNet, etc. These Neural Networks are trained on datasets like ImageNet which is a large dataset of images with a thousand classes. Initial layers of these networks can be frozen and the last couple of layers can be tuned according to our requirements and get the required number of outcome classes.

Researches have used this technique for skin cancer classification due to its reasonable reliability, low requirement of computational power, and decent output accuracy. To check how much better our model is as compared to the state of the art pre-trained model, we will compare the accuracies of our model and these pre-trained model side by side so as to get a better idea of the superiority of our model.

```
Layer (type)                 Output Shape              Param #
=================================================================
conv2d_4 (Conv2D)            (None, 27, 27, 64)        832
_________________________________________________________________
max_pooling2d_4 (MaxPooling2 (None, 13, 13, 64)        0
_________________________________________________________________
batch_normalization_4 (Batch (None, 13, 13, 64)        256
_________________________________________________________________
conv2d_5 (Conv2D)            (None, 12, 12, 512)       131584
_________________________________________________________________
max_pooling2d_5 (MaxPooling2 (None, 6, 6, 512)         0
_________________________________________________________________
batch_normalization_5 (Batch (None, 6, 6, 512)         2048
_________________________________________________________________
dropout_4 (Dropout)          (None, 6, 6, 512)         0
_________________________________________________________________
conv2d_6 (Conv2D)            (None, 5, 5, 1024)        2098176
_________________________________________________________________
max_pooling2d_6 (MaxPooling2 (None, 2, 2, 1024)        0
_________________________________________________________________
batch_normalization_6 (Batch (None, 2, 2, 1024)        4096
_________________________________________________________________
dropout_5 (Dropout)          (None, 2, 2, 1024)        0
_________________________________________________________________
conv2d_7 (Conv2D)            (None, 2, 2, 1024)        1049600
_________________________________________________________________
max_pooling2d_7 (MaxPooling2 (None, 2, 2, 1024)        0
_________________________________________________________________
batch_normalization_7 (Batch (None, 2, 2, 1024)        4096
_________________________________________________________________
dropout_6 (Dropout)          (None, 2, 2, 1024)        0
_________________________________________________________________
flatten_1 (Flatten)          (None, 4096)              0
_________________________________________________________________
dense_2 (Dense)              (None, 256)               1048832
_________________________________________________________________
dropout_7 (Dropout)          (None, 256)               0
_________________________________________________________________
dense_3 (Dense)              (None, 7)                 1799
=================================================================
```

*Figure 6. Model Summary of the CNN we developed.*

# 4. Training

The dataset used consists of 10015 images. The dataset is split into training and validation sets in 90% and 10% ratio. This leads to a model being trained in 9013 images and validated on 1002 images of the dataset. The model is trained with images resized to shape 28x28.

Now, the loss function used here while training is Categorical Cross-Entropy Loss Function. The reason behind using a Categorical Cross-Entropy Loss Function is that we have multiple output classes every image belongs only to one of the seven skin cancer types. Applying class weights to this imbalanced dataset is the most important thing to look for to achieve high accuracy. In the dataset used, Melanocytic nevi comprise most of the dataset. To counter fit this, the weight of Melanocytic class has been set to 0.5 and for all other classes, it has been set to 1. This will bring a drastic jump in final validation accuracy and will give a better classification report. The formula below represents general calculation for the Categorical Cross-Entropy Loss Function.

$$J = -\frac{1}{N}\left(\sum_{i=1}^{N} \mathbf{y_i} \cdot \log(\mathbf{\hat{y}_i})\right)$$

The choice of a perfect optimizer is very important. A wide range of options are available when it comes to optimizers such as Gradient Descent, Momentum, Nesterov, Adagrad, RMSProp, and the list goes on. But the best is the Adam optimizer which stands for Adaptive Moment Estimation. Adam is a combination of Adagrad, which works well on sparse gradients and RMSprop. Adam has β1 and β2 as hyper-parameters. This hyper-parameters β1 and β2 control the exponential decay rates of moving averages. For training our model we have set β1 = 0.9 and β2 = 0.999.

Having a proper learning rate during training is very important because it will determine how refined and accurate our final model will be. In this training, we start at a learning rate of 0.001 and it is decayed by a factor of 10 each time the validation loss plateaus after an epoch. We pick the model with the lowest validation loss in this process.

The complete dataset cannot be trained at once, we need to divide the dataset into the number of batches and train the model on that batch and move to another batch of images, and so on. Here the batch size used is 90 images. The model will train on 90 images and move to the other 90 images and this way will complete the whole epoch.

# 5. Results

The results of the model we built were satisfying. The precision, recall, and f1 score of each class was impressive enough to outshine the models based on Transfer Learning. The macro average and weighted average are also impressive enough. The table below shows a detailed report of the model we proposed.

| CLASS | PRECISSION | RECALL | F1 SCORE |
|---|---|---|---|
| 0 | 0.69 | 0.76 | 0.72 |
| 1 | 0.85 | 0.78 | 0.82 |
| 2 | 0.74 | 0.81 | 0.77 |
| 3 | 0.75 | 0.50 | 0.60 |
| 4 | 0.95 | 0.95 | 0.95 |
| 5 | 1.0 | 0.71 | 0.83 |
| 6 | 0.79 | 0.77 | 0.78 |
| **Macro average** | 0.82 | 0.76 | 0.78 |
| **Weighted average** | 0.89 | 0.89 | 0.89 |

| | |
|---|---|
| **Validation Accuracy** | 0.89 |

The Training accuracy achieved is 93% while the validation accuracy achieved is 89%. Figure 6 shows the graph of Training and Validation accuracy plotted against the epoch count at that particular instance.

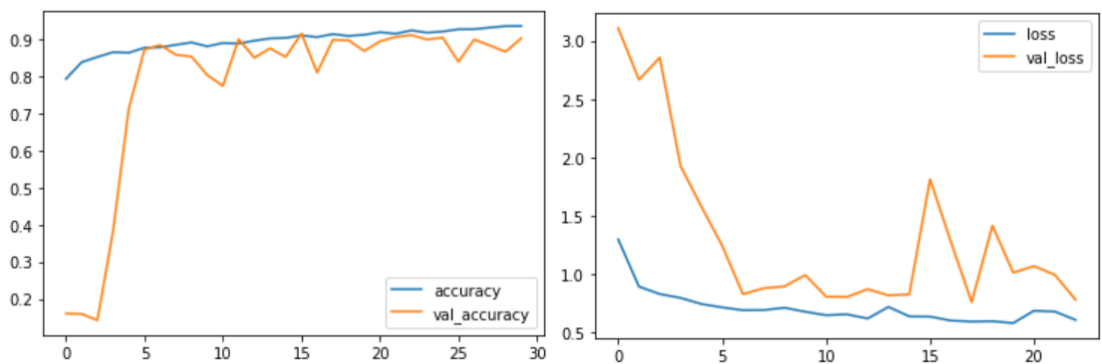

*Figure 7. Shows Training and Validation accuracy against epoch. On the other hand, it shows Training and Validation loss against epoch.*

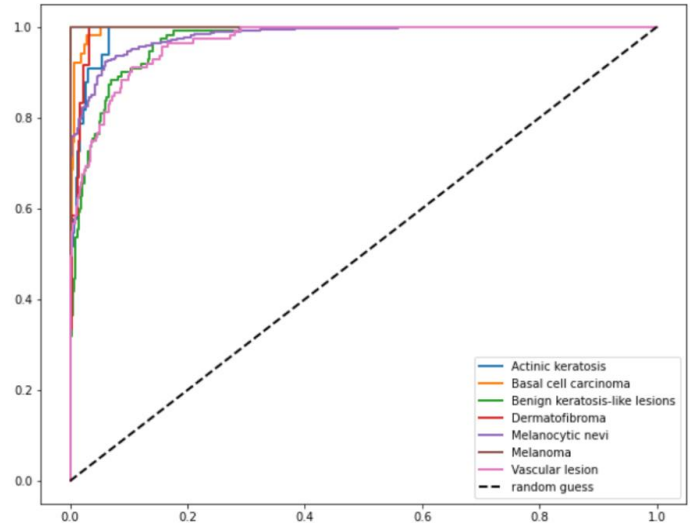

*Figure 8. ROC curve for the above performed training.*

# 6. Comparison

To see how our model has performed as compared to the different pre-trained models which use transfer learning, we can see the below table. It clearly shows that the model we built is far better or as good as ResNet50, InceptionV3, Vgg19, and MobileNetV2.

| MODEL | VALIDATION ACCURACY |
|---|---|
| Model Proposed in this Paper | **89%** |
| Vgg19 | 84% |
| InceptionV3 | 80% |
| ResNet50 | 79% |
| MobileNetV2 | 78% |

# 7. Conclusion

The Global Burden of Disease Study 2013 includes estimates of global morbidity and mortality due to skin disease. Skin conditions contributed 1.79% to the global burden of disease measured in DALYs from 306 diseases and injuries in 2013. Skin cancer is such a disease that needs to be identified quickly and in no time the proper treatment of the patient should be started, delay in treatment can lead to loss of life.

It can be seen that the CNN model we built is better than almost all the pre-trained models used for transfer learning like are ResNet50, InceptionV3, Vgg19, MobileNetV2, etc. This CNN model can be successfully used to identify skin cancer from any of these 7 classes with a very good accuracy. This CNN model can be relied upon to identify the presence of deadly cancer like Melanoma. This model has proved to be accurate and its capabilities can be put to good use by a dermatologist so as to identify deadly cancer within few seconds which can save precious lives of people all around the globe. We hope that this technology will be implemented to improve healthcare delivery and increase access to medical imaging expertise at remote locations of the world where access to skilled dermatologists and powerful testing facilities are limited.